\documentclass[letterpaper]{article}
\usepackage{aaai}
\usepackage{times}
\usepackage{helvet}
\usepackage{courier}
\usepackage[ruled,vlined]{algorithm2e}
\usepackage{algorithmic}
\usepackage{amsfonts}
\usepackage{amssymb}
\usepackage{graphicx}
\usepackage{url}
\usepackage{subfigure}
\usepackage{epstopdf}
\usepackage{multirow}
\usepackage{subfigure}
\usepackage{ifthen}
\newcommand{\FB}{{\sc FactorBase~}}
\newcommand{\TTuple}[1][0.0ex]{\vec{t}\hspace{#1}}

\newcommand{\UTuple}[1][0.0ex]{\vec{u}\hspace{#1}}

\newcommand{\VTuple}{\vec{v}}
\newcommand{\Mrange}[1]{\ifthenelse{\equal{#1}{T}}{\TTuple_m}{\ifthenelse{\equal{#1}{U}}{\UTuple_m}{\ifthenelse{\equal{#1}{V}}{\VTuple_m}{\mbox{UNKNOWN
TERM ID}}}}}
\newcommand{\Prange}[1]{\ifthenelse{\equal{#1}{T}}{\vec{t}_{pa}}{\ifthenelse{\equal{#1}{U}}{\vec{u}_{pa}}{\ifthenelse{\equal{#1}{V}}{\vec{v}_{pa}}{\mbox{UNKNOWN
TERM ID}}}}}

\newcommand{\GroundPrange}[1]{\ifthenelse{\equal{#1}{T}}{\vec{t}_{pa,\grounding'}}{\ifthenelse{\equal{#1}{U}}{\vec{u}_{pa,\grounding'}}{\ifthenelse{\equal{#1}{V}}{\vec{v}_{pa,\grounding'}}{\mbox{UNKNOWN
TERM ID}}}}}

\newcommand{\grounding}{\gamma}

\newcommand{\RVD}{{\it VDB }}

\usepackage{graphicx}
\usepackage{subfigure}

\usepackage{alltt}
\graphicspath{{../../}{figures/}}

\begin{document}

\title{SQL for SRL: Structure Learning Inside a Database System\\Position Paper}
\author{\\
Oliver Schulte \and Zhensong Qian\\
{oschulte,zqian@sfu.ca}\\
\\ School of Computing Science\\ Simon Fraser University\\Vancouver-Burnaby, Canada
}
\maketitle
\begin{abstract} 
The position we advocate in this paper is that relational algebra can provide a unified language for both representing and computing with statistical-relational objects, much as linear algebra does for traditional single-table machine learning. Relational algebra is implemented in the Structured Query Language (SQL), which is the basis of relational database management systems. To support our position, we have developed the \FB system, which uses SQL as a high-level scripting language for statistical-relational learning of a graphical model structure. The design philosophy of \FB is to manage statistical models as first-class citizens inside a database. 
Our implementation shows how our SQL constructs in \FB facilitate fast, modular, and reliable program development. Empirical evidence from six benchmark databases indicates that leveraging database system capabilities  achieves scalable model structure learning.
\end{abstract}

\section{Introduction} 
\begin{figure*}[!t]
\begin{center}
\resizebox{1\textwidth}{!}{
\includegraphics
%[width=0.8\textwidth]
{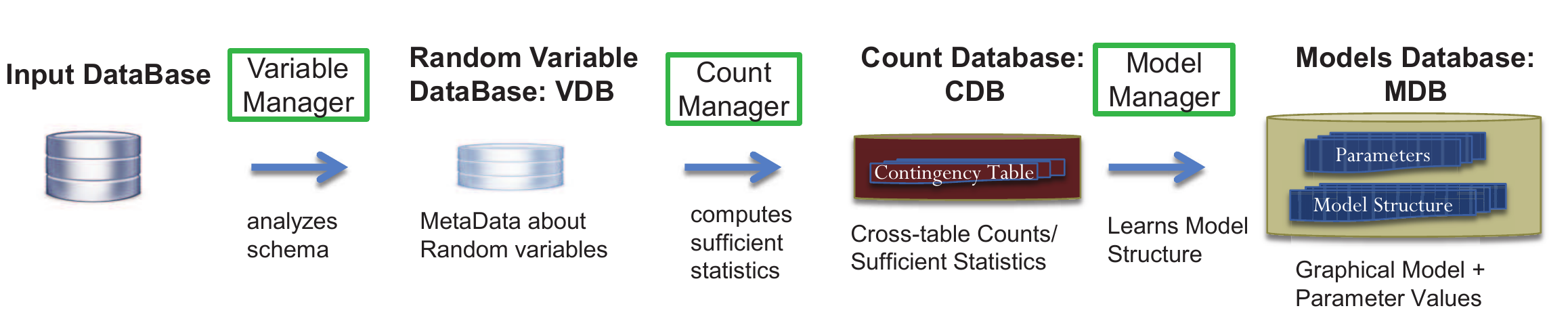}
}
\caption{System Flow. All statistical objects are stored as first-class citizens in a DBMS. Objects on the left of an arrow are utilized for constructing objects on the right. Statistical objects are constructed and managed by different modules (boxes). 
\label{fig:architecture}}
\end{center}
\end{figure*}

The statistical analysis of structured data requires structured machine learning models. Database researchers have developed a system architecture where statistical models are stored as first-class citizens inside a relational database management system (RDBMS) \cite{Wang2008,Niu2011}. The goal is to seamlessly integrate query processing and statistical-relational inference {\em inside the database}, rather than invoking external  procedures. 
These systems focus  on inference {\em given} a statistical-relational model, not on {\em learning} the model from the data stored in the RDBMS. 
We describe the \FB system that complements the in-database probabilistic inference systems with an in-database probabilistic model learning system. The name \FB indicates that our system supports learning a set of (par)-factors for a log-linear multi-relational model, typically represented in a graphical model~\cite{Kimmig2015}. 

There are several previous systems that leverage RDBMS support for learning \cite{MADlib_VLDB_2012,MLbase_ICDR_2013,Deshpande2006}, but they apply to traditional learning where the  data are represented in a {\em single} table or data matrix. The novel contribution of \FB is supporting graphical model learning for {\em multi-relational} data stored in different interrelated tables. The Sindbad system~\cite{Wicker2010} provides support for some multi-relational knowledge discovery tasks in an inductive database, but not for graphical model construction. Multi-relational graphical model construction raises several special challenges, such as:
\begin{enumerate}
\item A description language for specifying metadata about structured random variables.
\item Efficient mechanisms for constructing, storing, and transforming complex statistical objects, such as cross-table sufficient statistics, parameter estimates, and model selection scores.
\item Computing model prediction scores for relational test instances. 
\end{enumerate}

\FB applies SQL as a scripting language to implement database services that provide these capabilities. While \FB provides a good solution for each of these system capabilities in isolation, the ease with which large complex statistical-relational objects can be integrated via SQL queries is a key feature.

\subsection{Evaluation} Our system is fully implemented and source code is available available on-line~\cite{bib:bbsite}. We summarize the evaluation of \FB on six benchmark databases. For each benchmark database, the system applies the learn-and-join algorithm, a state-of-the-art SRL algorithm that constructs a  statistical-relational Bayesian network model \cite{Schulte2012}. The learned Bayes net structure can be converted to a Markov Logic network structure or a set of clauses \cite{Khosravi2010}.
%without configuration by user. 
The same SQL scripts work for all benchmark databases, which demonstrates the generality of our approach.  

Our experiments show that \FB pushes the scalability boundary: Learning scales to databases with over $10^6$ records, compared to less than $10^5$ for previous systems. At the same time it is able to discover more complex cross-table correlations than previous SRL systems. The scalability improvement is mainly due to the efficient computation and caching of sufficient statistics supported by SQL. Our experiments focus on two key services  for an SRL client: (1) Computing and caching sufficient statistics, (2) computing model predictions on test instances. The system can handle as many as 15M sufficient statistics. 
%We show that the predictions of a log-linear model can be computed using the Natural Join operator. 
SQL facilitates block-prediction for a set of test instances, which leads to a 10 to 100-fold speedup compared to a simple loop over test instances.

\subsection{Benefits}
We advocate using SQL as a high-level scripting language for SRL, because of the following advantages.

\begin{enumerate}
\item Extensibility and modularity, which support rapid prototyping. SRL algorithm development can focus on statistical issues and rely on the RDBMS for data access and processing.
\item Increased scalability, in terms of both the size and the complexity of the statistical objects that can be handled.
\item Generality and portability: standardized database operations support ``out-of-the-box'' learning with a minimal need for user configuration.
\end{enumerate}

\section{System Overview} 

Our system design represents statistical objects as  relational tables, on a par with the original data tables, so that SQL can be used  to manage them. 
Figure~\ref{fig:architecture} represents key system components. The starting point is a multi-relational database containing the input data. 

\subsection{System Components}  

\subsubsection{The Schema Analyzer} The schema analyzer is an SQL script that queries the system catalog table to define a default set of relational random variables (par-RVs) for statistical analysis \cite{Kimmig2015}. 
%The random variables represent descriptive attributes of entities, relationships among then, and descriptive attributes of these relationships. 
The metadata include the domain of the par-RVs (possible values), and type information (possible arguments). The schema analyzer extracts metadata about the random variables from the database system catalog.
%, which includes the following information. 
%(1) The domain of the random variable. For discrete random variables, this is a finite set of possible values.
%(2) Pointers to the table and/or column in the original database that contains the data relevant to the random variables. 
The random variables and associated metadata are stored in the \textbf{random variable database} $\RVD$. We highlight two features of the $\RVD$ component. 

(i) The set of par-RVs and the associated metadata is constructed {\em automatically} from the input database. 
%In contrast, existing SRL systems require users to specify information about par-RVs and associated types, typically in a system-specific format. 
Thus \FB utilizes the data description resources of SQL to faciliate the ``setup task'' for relational learning \cite{Walker2010}.

(ii) 
%In \FB, the downstream learning algorithm and scripts query metadata only in the \RVD, not in the system catalog. 
Representing metadata explicitly offers two advantages. 
First, a user can easily edit the \RVD to customize the learning behavior, for instance by deleting irrelevant par-RVs. Second, it is possible to export metadata from other formats to the \RVD format.
%, then use \FB to perform structure learning. 
In this way \FB can serve as a structure learning backend to expressive specification languages for other relational models \cite{Guazzelli2009,Milch2005}.

%
%Statistical analysis is based on a set of random variables. Single-table data is basically self-describing with respect to random variables: each column header other than row id fields represents a random variable. In contrast, a set of data tables that represents heterogeneous multi-relational data needs to be augmented with metadata about which columns represent which entity class. Such metadata requires a {\em data description language} \cite{Ullman1982}; in SQL the relevant key concepts are primary and foreign keys. 
%%Previous work on multi-relational learning typically assumed that information about random variables for heterogeneous is specified by the user using a logical language. 
%A novel aspect of \FB is analyzing the RDBMS system catalog to translate metadata about primary and foreign keys into a specification of relational random variables. 
%
%The Schema Analyzer examines the information in the DB system catalog to define a default set of random variables for statistical analysis.  Since the system catalog is itself stored in tables, the default set of random variables can be constructed using SQL queries. Metainformation about the random variables is stored in the This database may be edited by the user to add further random variables of interest. Another possibility is for a machine learning application to create the $\RVD$ database, for example based on metainformation specified in another format.
%

\subsubsection{The Count Manager}

A key service for statistical-relational learning is counting how many times a given relational pattern (par-RV) is instantiated in the data. Such counts are known as {\em sufficient statistics}. 
Accessing sufficient statistics is often the main scalability bottleneck. The access patterns of a model search procedure are inherently sequential and random \cite{Niu2011}, and therefore it is important to cache sufficient statistics.  Caching is even more important if the data is stored on disk in an RDBMS, rather than in main-memory. 
There are several reasons for employing an RDBMS for gathering sufficient statistics. 
(1) The machine learning application saves expensive data transfer by executing count operations in the database server space rather than local main memory. (2) SQL optimizations for aggregate functions such as SUM and COUNT can be leveraged. (3) Sufficient statistics can be stored in the RDBMS. For many datasets, the number of sufficient statistics runs in the millions and is too big for main memory.  
%
%An RDBMS provides disk storage and fast access for large numbers of sufficient statistics. 
A novel aspect of \FB is managing {\em multi-relational sufficient statistics} that combine information {\em across} different tables in the relational database. This requires combining SQL aggregate functions with table joins \cite{Qian2014a}.

 %Figure~\ref{fig:metaquery-concept} illustrates this design. 
%
%
%A novel aspect of \FB is storing sufficient statistics in contingency tables defined by database views. These views are defined by SQL queries, which are constructed using the novel concept of an SQL metaquery. An SQL metaquery takes as input the metainformation about random variables, and builds a set of tables that represent a view definition. 

\subsubsection{The Model Manager} 

The Model Manager supports the construction and querying of large structured statistical models, which are stored in the \textbf{Model Database} MDB. Services provided by the Model Manager include the following. (1) Compute parameter estimates for the model using the sufficient statistics in the Count Database.  (2) Computing model characteristics such as the number of parameters or degrees of freedom in a model. (3) Computing a model selection score that quantifies how well the model fits the multi-relational data. 
Model selection scores are usually functions of the number of parameters and the parameter estimates.  By employing the SQL view mechanism, parameter estimates and model selection scores are updated automatically during the model search.

\bibliographystyle{aaai}
\bibliography{starai15} 

\end{document}